\IfFileExists{eccv-paper/llncs.cls}{%
  \documentclass[runningheads]{eccv-paper/llncs}%
}{%
  \documentclass[runningheads]{llncs}%
}

\IfFileExists{eccv-paper/eccv.sty}{%
  \usepackage[final,year=2026]{eccv-paper/eccv}%
}{%
  \usepackage[final,year=2026]{eccv}%
}
\IfFileExists{eccv-paper/eccvabbrv.sty}{%
  \usepackage{eccv-paper/eccvabbrv}%
}{%
  \usepackage{eccvabbrv}%
}

\usepackage{graphicx}
\graphicspath{{images/}{eccv-paper/images/}}
\usepackage{float}
\usepackage{amsmath,amssymb}
\usepackage{booktabs}
\usepackage{multirow}
\usepackage{xcolor}
\usepackage{algorithm}
\usepackage{algpseudocode}
\usepackage{tikz}
\usepackage{soul}
\usetikzlibrary{arrows.meta,positioning,fit,backgrounds,calc,shapes.geometric}
\usepackage[pagebackref,breaklinks,colorlinks,citecolor=eccvblue]{hyperref}

\newcommand{\paperinput}[1]{\IfFileExists{eccv-paper/#1}{\input{eccv-paper/#1}}{\input{#1}}}

\begin{document}

\title{RedLight-VLA: Models for traffic-rule grounding and behavioral emphasis in driving policies}
\titlerunning{RedLight VLA - What to Represent and What to Emphasize}

\author{%
Bala Murali Manoghar Sai Sudhakar\inst{1} \and
Sourab Bapu Sridhar\inst{2} \and
Sandipan Das\inst{3} \and
Rahul Ahuja\inst{1} \and
Meda Lazar\inst{4} \and
Ashish Garg\inst{3} \and
Pratik Likhar\inst{3} \and
Senthil Yogamani\inst{1}%
}
\authorrunning{B. M. Manoghar Sai Sudhakar et al.}

\institute{%
Qualcomm Technologies, Inc., USA\\
\email{\{balasudh,rahahu,syogaman\}@qti.qualcomm.com}%
\and
Qualcomm Auto Ltd Sweden Filial, Sweden\\
\email{soursrid@qti.qualcomm.com}%
\and
Qualcomm India Private Limited, India\\
\email{\{imsandi,ashgarg,plikhar\}@qti.qualcomm.com}%
\and
Arriver System Software S.r.l., Romania\\
\email{mlazar@qti.qualcomm.com}%
}
\date{}

\maketitle

\begin{abstract}

Behavior-cloned Vision-Language-Action (VLA) driving policies struggle with
rare rule-governed maneuvers at signalized intersections. Braking and
launching examples contribute little to averaged trajectory loss, while fused
representations lack explicit supervision for the governing traffic-light and
stop-line state. We present RedLight-VLA, a training
objective that uses expert futures and automatically generated perception
targets without additional manual rule annotation. First, trajectory-derived
behavioral reweighting (BR) emphasizes rare deceleration and acceleration using
rotation-invariant longitudinal dynamics and a scale-preserving reduction that
exactly recovers the baseline when disabled. Second, parallel auxiliary (AUX) heads
ground traffic-light and stop-line state in continuous post-fusion rule tokens,
without autoregressive language generation or changes to the trajectory
decoder. We evaluate on a curated set of
$20\,$s sequences with a $5\,$s prediction horizon. Controlled variants share the same
backbone, training data, decoder, and evaluation population. Against an
otherwise identical VLA
baseline, RedLight-VLA reduces red-light stop-line overshoot from $7.3\%$ to
$6.8\%$, reduces stop-line velocity error by $12.7\%$, and improves $3\,$s
traffic-light-sliced ADE/FDE from $0.274/0.964\,$m to $0.247/0.897\,$m.
Green-light false stops increase from $3.2\%$ to $3.9\%$; however, combining BR with AUX supervision mitigates the larger increase observed for AUX alone ($4.0\%$). The combined model also improves non-traffic-light
ADE/FDE from $0.268/0.956\,$m to $0.241/0.876\,$m and outperforms either
mechanism alone on all four sliced displacement measures.

\end{abstract}


\section{Introduction}
\label{sec:intro}

Imitation learning (IL) is the standard approach for end-to-end
driving: a model learns to reproduce the expert's future trajectory
from logged demonstrations~\cite{bojarski2016end,bansal2019chauffeurnet, joseph2021autonomous}.
Its main appeal is scale, but that scale hides a structural weakness.
Most frames in a driving log show smooth, near-constant-velocity lane
keeping. The behaviors that matter most for safety and comfort are
rare: a controlled stop behind a lead vehicle, a launch from a green
light, or a brake for a cut-in. This causes challenges in creating a balanced training dataset~\cite{uricar2019challenges}. Recent Vision-Language-Action (VLA)
driving models~\cite{brohan2023rt2,kim2024openvla,%
tian2024drivevlm} inherit the same objective and the same weakness.
Two complementary limitations cause it: the objective does not
emphasize rare but safety-critical behaviors, and the representation
lacks explicit supervision of rule-governed events.

\emph{(i) Rare kinematics.} Under a mean-reduced regression
objective, if hard brakes make up (say) $2\%$ of frames, they
contribute a commensurately small share of the
gradient, even though a
missed brake is far more costly than a slightly imperfect cruise.
Two asymmetric failure modes recur. The first is \emph{under-braking},
that is, sluggish or late deceleration. The second is the ``stops
well but cannot resume'' problem, in which the rare launch-from-rest
transition is under-fit. Both are longitudinal, both live in the
tail of the acceleration distribution, and both are invisible to
aggregate displacement error.

\emph{(ii) Rare rule-governed events.} Signalized intersections,
with their lights, stop lines, and stop-and-go transitions, are
exactly where behavior cloning is thinnest and mistakes are most
costly. Trajectory supervision alone does not explicitly identify the
traffic-light and stop-line state governing the correct action, while
expressing that state as text couples rule grounding to tokenizer-dependent
autoregressive decoding. We ask whether a VLA policy can learn both the
rare longitudinal tail and the rule state without additional manual rule
annotation or language generation at deployment.

\noindent\textbf{Our key idea:} the required signals already exist in the
training stack. The expert future identifies \emph{which examples to
emphasize}, motivating \textbf{behavioral reweighting}; automatically
generated perception targets identify \emph{what information to
represent}, motivating \textbf{auxiliary rule-token supervision} of the
fused VLA token space. Neither requires new sensors or manual rule labels.

\begin{figure}[!ht]
\centering
\includegraphics[width=0.95\linewidth]{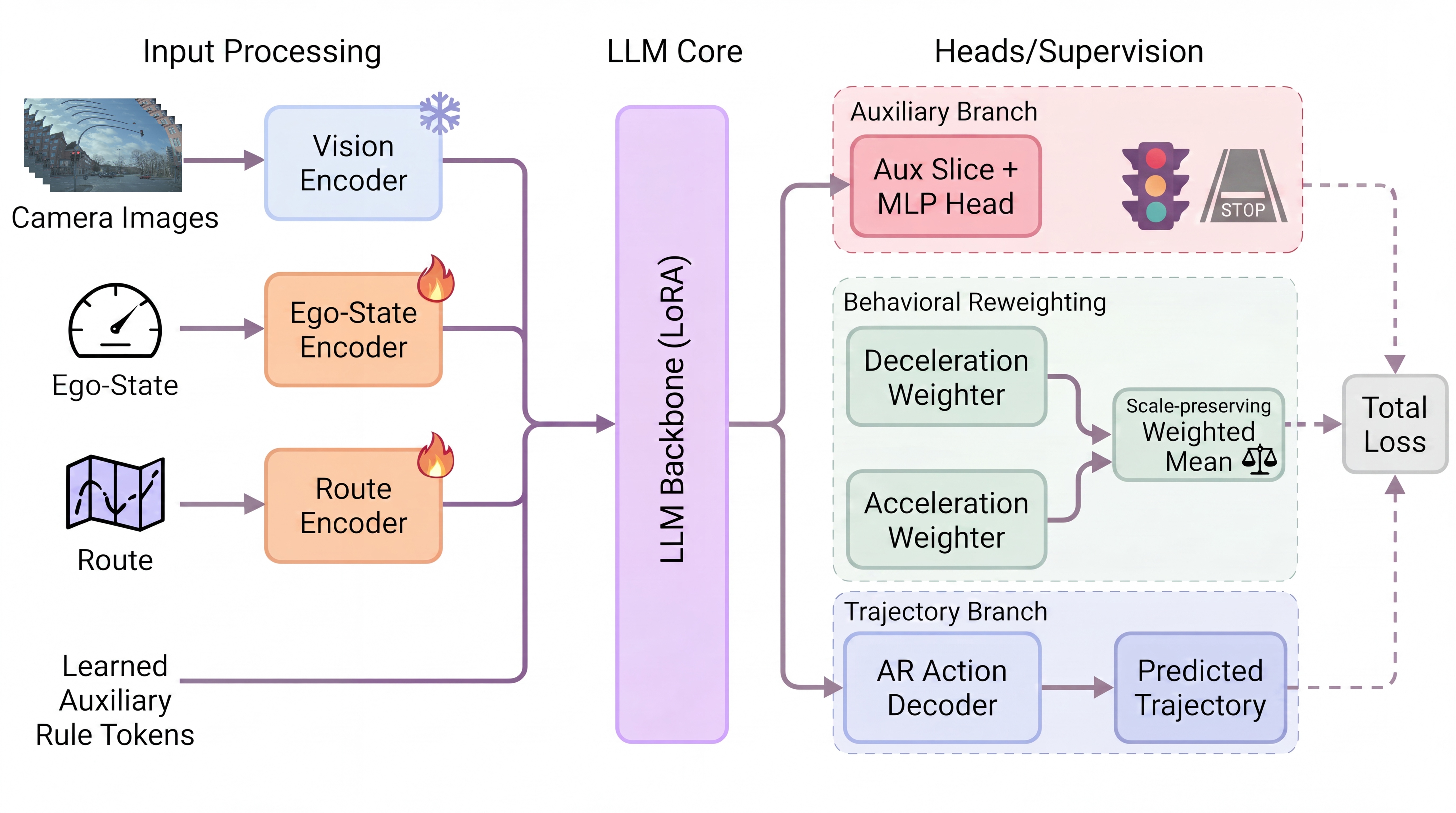}
\caption{\textbf{Overview of the rule-token and behavioral-reweighting pipeline.} A VLA trajectory predictor fuses vision, ego-state, and route tokens through an LLM backbone and action decoder to predict a delta trajectory. \textbf{Behavioral reweighting (green)} scores the ground-truth trajectory's longitudinal dynamics and combines deceleration/acceleration weights into a per-sample weight for a scale-preserving weighted mean. \textbf{Auxiliary rule-token supervision (red)} appends learned placeholder tokens to the token bus; an MLP head reads the fused slots and is supervised against the perception stack's traffic-light and stop-line outputs. Disabling both paths recovers the standard masked-mean objective.}
\label{fig:system}
\end{figure}

The two mechanisms are deliberately decoupled: both act at the loss
layer, neither modifies the trajectory head, and each can be enabled
independently. Our 2$\times$2 experimental design
(Section~\ref{sec:protocol}) tests whether their benefits compound when
the mechanisms are combined.
For the representation mechanism, we depart from the common recipe of
an auxiliary head on a shared backbone feature. We instead reserve a
few tokens on the model's own token bus and read the rule state from
them \emph{after} language-model fusion, so the supervision lands on
the same fused context the trajectory decoder consumes, rather than on
an early feature that may be diluted before fusion
(Fig.~\ref{fig:system}).

\noindent\textbf{Our contributions} are as follows:
\begin{enumerate}
  \item \textbf{A framework without additional manual rule annotation}
    that couples two signals
    the stack already carries (label-derived kinematics and
    perception derived rule state) to two independently toggled
    loss-layer mechanisms, together with a joint-coverage analysis of
    how the kinematic and rule tails overlap in driving logs
    (Section~\ref{sec:method}).
  \item \textbf{Trajectory-derived behavioral reweighting} from
    rotation-invariant longitudinal dynamics of the expert future, via
    two composite weighters and a \emph{scale-preserving} weighted
    mean that keeps loss hyperparameters calibrated and recovers the
    baseline objective exactly when disabled
    (Section~\ref{sec:reweight}).
  \item \textbf{Supervised rule registers}: reserved post-fusion
    tokens with parallel multi-layer perceptron (MLP) heads that
    ground traffic-light and stop-line state in the fused VLA
    representation without language decoding or changes to the
    trajectory path
    (Section~\ref{sec:auxtokens}).
  \item \textbf{An open-loop rule-compliance suite} for real logs,
    covering red-light stop-line overshoot rate, green-light
    false-stop rate, and stop-line velocity error, all computable from
    perception signals already in the stack and complementing weakly
    diagnostic displacement metrics (Section~\ref{sec:protocol}).
\end{enumerate}

\vspace{-2mm}
\section{Related Work}
\vspace{-1mm}
\label{sec:related}

\paragraph{Imitation learning for driving.}
Behavior cloning underpins a wide range of learned driving methods, from end-to-end pixel-to-control models~\cite{bojarski2016end}, through post-perception trajectory predictors~\cite{bansal2019chauffeurnet,gao2020vectornet,liang2020lanegcn,%
shi2022mtr,nayakanti2023wayformer}, to planning-oriented and language-conditioned
policies~\cite{hu2023uniad,tian2024drivevlm,brohan2023rt2,kim2024openvla}. Prior
work addresses the covariate shift that appears during rollout through
DAgger~\cite{ross2011dagger} and trajectory-perturbation
augmentation~\cite{bansal2019chauffeurnet}, and
Codevilla~\etal~\cite{codevilla2019exploring} document the limits of pure
behavior cloning well. We target a different
failure mode: the imbalance of the demonstration distribution and the absence of
rule-level supervision. Our approach does not depend on how the trajectory is
decoded, and it complements the remedies for distribution shift.

\paragraph{Reweighting, mining, and long-tail learning.}
Countering imbalance by reweighting has a long history:  effective-number
reweighting~\cite{cui2019classbalanced}, focal loss that down-weights easy
examples~\cite{lin2017focal}, label-distribution-aware margins~\cite{cao2019ldam},
hard-example mining~\cite{shrivastava2016ohem}, importance
sampling~\cite{katharopoulos2018importance}, and learned per-task or per-sample
weights such as uncertainty~\cite{kendall2018multitask} or gradient-norm
balancing~\cite{chen2018gradnorm}. These methods either require class labels, tie
the weight to a non-stationary optimizer state, or add learned parameters. In contrast, we reweight a continuous trajectory regression objective using an \emph{a~priori} behavioral property of the expert future. The weight is label-derived, fixed, interpretable, and parameter-free, and it composes with prior approaches rather than supplanting them. For driving, the long tail is more often addressed at the
data level, either through decoupled rebalancing~\cite{kang2020decoupling} or
through generated safety-critical
scenarios~\cite{wang2021advsim,rempe2022strive}. Our method offers a
complementary mechanism at the loss level.

\paragraph{Auxiliary tasks for driving policies.}
Auxiliary supervision is a standard way to shape representations in
reinforcement learning and perception~\cite{jaderberg2017reinforcement}, and
planning-oriented stacks supervise intermediate perception and prediction
tasks ~\cite{hu2023uniad}. Most methods attach auxiliary heads like depth estimation~\cite{kumar2018near} to a shared backbone.
Concept Bottleneck Models~\cite{koh2020concept} instead route the output through
a concept layer, which reduces primary-task accuracy. Our approach differs on
both counts: we read the rule state from reserved tokens \emph{after}
language-model fusion, not from an early backbone feature, and our lightweight
head leaves the prediction path untouched. Framed differently, our reserved slots act as \emph{supervised registers}: learnable placeholders appended to the sequence, filled by attention, and read after fusion. Unlike prior registers, which emerge without supervision, their semantics are explicitly grounded by the label. In contrast to the unsupervised registers of Darcet~\etal~\cite{darcet2024registers}, a perception-derived target dictates what our slots must encode. They are structurally similar to the learned query tokens of Q-Former~\cite{li2023blip2}
and OmniDrive~\cite{wang2024omnidrive}, but our tokens are supervised against
symbolic rule state.

\paragraph{Traffic-rule-aware driving and how it is measured.}
Two lines of work study rule awareness at signalized intersections. In the
\emph{closed-loop CARLA simulator}~\cite{carla_leaderboard} line, a perception-side head predicts traffic-light
state and feeds a rule-based safety controller
(InterFuser~\cite{shao2022interfuser}, ReasonNet~\cite{shao2023reasonnet}), and
results are scored with Driving Score and red-light infraction
rate~\cite{carla_leaderboard,wu2022tcp}. LMDrive~\cite{shao2024lmdrive} learns a
traffic-light token, but only during \emph{pre}-LLM vision-encoder pre-training,
and it discards the associated head afterward, so it never reads the rule
signal from the fused sequence. In the \emph{language/visual question answering (VQA)} line,
DriveLM~\cite{sima2024drivelm} answers graph-structured questions, and recent
VLA planners reason about rules in text or report closed-loop scores rather than
rule-sliced displacement (SimLingo~\cite{renz2025simlingo},
ORION~\cite{fu2025orion}, Alpamayo-R1~\cite{wang2025alpamayo}).
D2-TPred~\cite{zhang2022d2tpred} conditions on lights but reports only aggregate
displacement. The closest work in spirit,
EMMA~\cite{hwang2024emma}, represents driving outputs as \emph{decoded text
through the language head}, coupling prediction to autoregressive decoding and its
tokenizer. We instead supervise \emph{continuous reserved tokens with parallel
MLP heads}, adding no decode steps and no tokenizer coupling. We evaluate with
traffic-light-sliced ADE/FDE and rule-compliance metrics
(Section~\ref{sec:protocol}) on real-world logs.


\vspace{-2mm}
\section{Method}
\vspace{-2mm}
\label{sec:method}

Our method adds two pieces to a standard VLA trajectory trainer. The first,
\emph{behavioral reweighting} (Section~\ref{sec:reweight}), makes rare
maneuvers such as hard brakes and launches from rest count more in the
training loss. The second, \emph{auxiliary rule-token supervision}
(Section~\ref{sec:auxtokens}), pushes the model's internal representation to
explicitly contain the traffic-light and stop-line state. Both act only on
the loss; the network that predicts trajectories is unchanged. We first
describe the base model and notation (Section~\ref{sec:setup}).
\vspace{-2mm}
\subsection{Setup and notation}
\label{sec:setup}

We build on a VLA trajectory predictor (Fig.~\ref{fig:system}). A vision
encoder turns a short history of front-camera frames (five frames,
${\approx}0.5$\,s) into visual tokens. Further tokens encode the past ego
motion and the navigation route or standard-definition (SD) map. All of
these tokens form one input sequence, which we call the \emph{token bus}. A
small (${\sim}0.6$B-parameter) open-weight language model fuses this
sequence in one context pass per planning query. A separate autoregressive
action decoder, adopted from SMART~\cite{wu2024smart}, then predicts the future
trajectory over a fixed horizon. The vision encoder
and the language model are adapted with Low-Rank Adaptation
(LoRA) ~\cite{hu2022lora}. When
enabled, the auxiliary rule tokens of Section~\ref{sec:auxtokens} join the
same token bus.


For each training sample $i$, the label is the trajectory the expert
actually drove, stored as per-step displacements
$\boldsymbol{\Delta}_i \in \mathbb{R}^{T\times 2}$: how far the vehicle
moved in $(x,y)$ during step $t$, with $T=50$ and $\Delta t=0.1$\,s. A
validity mask $\mathbf{m}_i \in \{0,1\}^{T}$ marks padded timesteps. For
every frame, the perception stack additionally reports the ego-relevant
traffic-light state and the nearest stop-line geometry;
Section~\ref{sec:auxtokens} uses these as automatically generated targets.
The action decoder is trained with a per-step displacement
regression loss and a classification loss over discretized
longitudinal-acceleration and heading bins, following standard
trajectory-prediction practice~\cite{nayakanti2023wayformer,shi2022mtr},
reduced by the masked mean of Eq.~\eqref{eq:weighted-mean}. The behavioral
weights of Section~\ref{sec:reweight} apply to these terms through the same
reduction, and the scale-preserving property is what keeps the loss scale
unchanged when reweighting is enabled.

\vspace{-2mm}
\subsection{Behavioral reweighting}
\label{sec:reweight}

Most training frames show easy, near-constant motion, so rare maneuvers
contribute almost nothing to an averaged loss. Behavioral reweighting
counters this by assigning every sample a weight
$w_i \in [1, w_{\max}]$: ordinary samples keep weight $1$, and rare braking
or launching samples receive more. The weight is computed from the
ground-truth future itself, so no extra labels are needed. We call a rule
that assigns these weights a \emph{weighter}. A weighter is a pure,
stateless function of the batch, which makes weighters
composable and safe under distributed training; when nothing rare is
detected, a weighter returns $1$.

\paragraph{Step 1: read the dynamics off the label.}
From the per-step displacements, we compute how fast the expert was moving
and how strongly it was accelerating:
\begin{align}
  s_{i,t}   &= \frac{\lVert \boldsymbol{\Delta}_{i,t}\rVert_2}{\Delta t},
              && t = 1,\dots,T, \label{eq:speed}\\[2pt]
  a_{i,t}   &= \frac{s_{i,t+1}-s_{i,t}}{\Delta t},
              && t = 1,\dots,T-1. \label{eq:accel}
\end{align}
Here $s_{i,t}$ is the distance traveled in step $t$ divided by the step
duration, and $a_{i,t}$ is its per-second change;
$s_{i,0}=\sqrt{v_{x,i,0}^2+v_{y,i,0}^2}$ is the current ego speed. We use
the \emph{length} of the displacement vector rather than its forward
component, so $s$ is unaffected by turning: the same thresholds mean the
same thing on straights and through arbitrary turn geometries.

\paragraph{Step 2: triggers.}
A \emph{trigger} is a yes/no test on these dynamics, paired with a weight.
The \textbf{deceleration weighter (DCCL)} targets braking with three
triggers. (i)~\emph{Sustained deceleration}: the future contains at least
$0.5$\,s of deceleration stronger than $-1.5\,\text{m/s}^2$.
(ii)~\emph{Hard brake}: any single step decelerates harder than
$-3.0\,\text{m/s}^2$. (iii)~\emph{General slowing}: the ego is moving at
the start of the horizon ($s_{i,0}\geq v_{\text{stop}}$, with
$v_{\text{stop}}=1.0$\,m/s) and the future
minimum speed falls more than $30\%$ below the initial speed,
$\min_t s_{i,t}/\max(s_{i,0},\epsilon)<1-\gamma$ with $\gamma{=}0.30$ and
$\epsilon{=}1$ guarding against division by zero; the moving gate keeps
already-stationary frames from firing this trigger.

The \textbf{acceleration weighter (ACCL)} targets the opposite failure, a
model that stops well but never resumes. (i)~\emph{Launch from rest}: the
ego is currently stopped ($s_{i,0}<v_{\text{stop}}$) but the
future sustains acceleration above $\theta_{\text{launch}}$ for at least
$\tau_{\text{launch}}$, where these denote the launch-acceleration threshold
and its minimum consecutive duration. (ii)~\emph{Hard acceleration}: any step above
$2.5\,\text{m/s}^2$. (iii)~\emph{Speed-up}: the future maximum speed
exceeds the initial speed by more than $2.0$\,m/s. Trigger weights are fixed
before training from their empirical frequencies in the training split, with
rarer events receiving larger weights.
\vspace{-2mm}
\paragraph{Step 3: combine the triggers.}
Triggers within one weighter can overlap, so a sample takes the largest
active weight (a $\max$; nothing is double-counted). Across weighters,
the phenomena are distinct, so the two outputs multiply, clipped to
$[1,w_{\max}]$ with $w_{\max}=15$:
\begin{equation}
  w_i \;=\; \mathrm{clip}\!\Big(\textstyle\prod_{k\in\{\mathrm{DCCL},\mathrm{ACCL}\}} w^{(k)}_i,\; 1,\; w_{\max}\Big).
  \label{eq:composite}
\end{equation}
A sample that is rare along both axes, such as a stop followed by a
launch at the same intersection, is emphasized more than one rare along a
single axis. Every factor is at least $1$, so no sample ever counts less
than baseline, and a future axis of importance enters as one more factor
without touching the loss layer.

\paragraph{Step 4: apply the weights without breaking the loss scale.}
There is one pitfall in using the weights. If each sample's loss were
simply multiplied by its weight and averaged, the loss would grow with
the weights, which silently acts like a larger learning rate and detunes
every other loss coefficient. We instead rescale the weights so that they
average to $1$ over the valid elements of the batch, and only then
average. Given a per-element loss $\ell$, its mask $m$, and weights $w$, define
the reduction $\mathcal{R}_w$ as
\begin{equation}
  \bar{w}_{i} = w_i \cdot
    \frac{\sum_{j,\tau} m_{j,\tau}}{\sum_{j,\tau} m_{j,\tau}\, w_j},
  \qquad
  \mathcal{R}_w(\ell;m) = \frac{\sum_{i,t} m_{i,t}\,\bar{w}_{i}\,\ell_{i,t}}
                     {\sum_{i,t} m_{i,t}},
  \label{eq:weighted-mean}
\end{equation}
where the normalization is computed over the global batch via the
distributed all-reduce. Two properties follow. First, when every weight
is $1$, Eq.~\eqref{eq:weighted-mean} is exactly the ordinary masked mean,
so disabling reweighting recovers the baseline bit-for-bit. Second, the
reduction is invariant to any global rescaling of the weights
($w \mapsto cw$ leaves $\mathcal{R}_w$ unchanged): the
weights redistribute gradient between samples without inflating the loss
scale, so no loss coefficient needs re-tuning. Computing the weights is
an $O(BT)$ pass over the labels, negligible next to a backbone forward
pass.
\subsection{Auxiliary rule-token supervision}
\label{sec:auxtokens}

Reweighting decides \emph{which} samples matter; it does not tell the
model \emph{what} a red light is. The second mechanism supplies that: we
make the fused representation explicitly predict the traffic-rule state,
using labels the perception stack already produces.

\paragraph{Signal.}
For every frame, the perception stack reports (a)~the ego-relevant
traffic-light state $y^{\mathrm{tl}}_i \in \{\text{none}, \text{green},
\text{red}\}$, and (b)~the nearest transverse stop line: whether one
exists, its signed longitudinal distance $\rho_i$ (m), and its
orientation as a unit normal $(\sin,\cos)$. A stop line counts only if it
is the nearest solid line inside the ego corridor ($\leq 6$\,m lateral
offset) and within $30^{\circ}$ of perpendicular to travel. Both signals
are byproducts of the running stack; nothing new is annotated.

\paragraph{Reserved tokens and heads.}
For each auxiliary task $k$, we append $n_k$ learned placeholder
tokens to the token bus: blank slots, identical for every sample,
analogous to action query tokens. They are inserted among the context
tokens, before the positions from which the action decoder reads, so the
fused states that condition the trajectory can attend to them. The LLM's
attention fills the slots with scene information during fusion; after the
LLM, we slice the hidden states at exactly these slots,
$\mathbf h_i^{(k)}\in\mathbb{R}^{n_k\times D}$, where
$k\in\{\mathrm{tl},\mathrm{sl}\}$ indexes the task and $D$ is the LLM hidden
dimension, and read them with a small MLP head (flatten, then
\texttt{Linear}--\texttt{GELU}--\texttt{LayerNorm}--\texttt{Linear}).
Because the head sees only the reserved slots, the model must write the
rule state into them, and because the slots live in the fused sequence
the trajectory decoder conditions on, that state is available to the
trajectory path as well.

We instantiate two independently switchable tasks, each with its own
tokens and head. The \emph{traffic-light relevance} task
($n_{\mathrm{tl}}=1$) predicts a presence logit $\hat p^{\mathrm{tl}}$
(is there an ego-relevant light?) and a color logit $\hat c^{\mathrm{tl}}$
(red vs.\ green). The \emph{stop-line} task ($n_{\mathrm{sl}}=2$)
predicts presence $\hat p^{\mathrm{sl}}$, distance $\hat\rho$, and
orientation $(\widehat{\sin},\widehat{\cos})$:
\begin{equation}
  \big(\hat p^{\mathrm{tl}}_i,\hat c^{\mathrm{tl}}_i\big)
     = \mathrm{MLP}_{\mathrm{tl}}\big(\mathbf h_i^{(\mathrm{tl})}\big),
  \qquad
  \big(\hat p^{\mathrm{sl}}_i,\hat\rho_i,\widehat{\sin}_i,\widehat{\cos}_i\big)
     = \mathrm{MLP}_{\mathrm{sl}}\big(\mathbf h_i^{(\mathrm{sl})}\big).
  \label{eq:tlhead}
\end{equation}
With both tasks enabled, the model carries three extra tokens and two
heads, negligible in parameters and FLOPs.

\paragraph{Losses.}
The traffic-light loss is a binary cross-entropy on presence, plus a
color cross-entropy applied only when a light is present
(presence-gated). The stop-line loss is a binary cross-entropy on a
\emph{capped-presence} target: a stop line counts as present only if it
also passes forward-distance and orientation caps ($\rho \leq 40$\,m,
within $30^{\circ}$ of transverse); lines too far or too skewed count as
absent. On the capped-present subset, we add a smooth-$L_1$ loss on the
normalized distance. For target $\mathbf u=(\sin,\cos)$ and prediction
$\hat{\mathbf u}=(\widehat{\sin},\widehat{\cos})$, we use
$\ell_{\mathrm{ori}}=\mathrm{SmoothL1}(\hat{\mathbf u},\mathbf u)$ and
$\ell_{\mathrm{circle}}=(\lVert\hat{\mathbf u}\rVert_2^2-1)^2$.
Task $k$'s loss is $\mathcal{L}^{(k)}_{\mathrm{aux}}=\sum_l
\lambda^{(k)}_l\ell^{(k)}_l$, where $l$ indexes presence/color for TL and
presence/distance/orientation/circle for stop lines, and
$\lambda^{(k)}_l\geq0$ is a fixed coefficient. The total
training objective combines the primary trajectory loss with the
auxiliary losses:
\begin{equation}
  \mathcal L=\mathcal L_{\mathrm{primary}}
  +\sum_{k\in\{\mathrm{tl},\mathrm{sl}\}}\mathcal L^{(k)}_{\mathrm{aux}}.
  \label{eq:total}
\end{equation}
If a batch contains no positive samples, the gated terms would drop out
of the computation graph and desynchronize the distributed all-reduce; a
zero-scaled reduction of the head output keeps them connected.


\section{Experiments}
\label{sec:protocol}

\paragraph{Dataset.}
We train and evaluate on a curated driving dataset. Table~\ref{tab:dataset_stats}
summarizes the train, validation, and test splits. The inputs are
front-wide imagery (5-frame history over ${\approx}0.5$\,s at
$480\times960$), a $1.6$\,s kinematic history
($[v_x,v_y,a_x,a_y]$ at $\Delta t{=}0.1$\,s), an SD map, calibration,
and goal points; the network predicts a fixed horizon
future trajectory. The auxiliary targets are read directly from existing
perception outputs, traffic-light tensors for light relevance and
road-line polylines for stop lines (Section~\ref{sec:auxtokens}), so no
additional manual annotation is introduced, though the targets inherit
upstream perception errors. A frame belongs to the \emph{TL slice} when
the perception stack resolves an ego-relevant light in
$\{\text{green},\text{red}\}$ or emits a qualifying transverse stop
line.

\begin{table}[!ht]
\centering
\caption{\textbf{Dataset scale and split composition.} Percentages are relative
to all frames in each split; parentheses report frame counts, with ``K'' denoting
thousands for the training split. ``Resolved TL'' is the red/green target used
by the auxiliary task, and ``Stop-line'' is the capped target used by the
stop-line head.}
\label{tab:dataset_stats}
\scriptsize
\setlength{\tabcolsep}{3pt}
\resizebox{\linewidth}{!}{%
\begin{tabular}{lrrrrrr}
\toprule
Split & Sequences & Frames & Any TL (raw) & Ego-rel. TL (raw) & Resolved TL (GT) & Stop-line (GT) \\
\midrule
Train & $235{,}985$ & $1{,}651{,}895$ & $40.66\%$ ($671$K) & $38.34\%$ ($633$K) & $25.70\%$ ($425$K) & $10.10\%$ ($167$K) \\
Val & $16{,}318$ & $114{,}226$ & $40.62\%$ ($46{,}395$) & $38.26\%$ ($43{,}703$) & $25.59\%$ ($29{,}236$) & $10.01\%$ ($11{,}429$) \\
Test & $16{,}404$ & $114{,}828$ & $40.69\%$ ($46{,}727$) & $38.43\%$ ($44{,}128$) & $25.81\%$ ($29{,}640$) & $10.19\%$ ($11{,}696$) \\
\bottomrule
\end{tabular}%
}
\end{table}

\paragraph{Base model.}
We evaluate inside the VLA trajectory predictor of Section~\ref{sec:setup},
applying reweighting to the trajectory-loss terms via
Eq.~\eqref{eq:weighted-mean}; the auxiliary
head reads reserved token slots after the LLM. Unless stated otherwise,
training uses AdamW ~\cite{loshchilov2019adamw} with bfloat16 precision and gradient clipping, on a
multi-GPU cluster of H100 GPUs. The learning rate warms up, holds for part
of training, then cosine-decays with batch-size scaling; LoRA ~\cite{hu2022lora} adapts the LLM q/k/v/o projections and the vision q/k/v/out projections.

\paragraph{Conditions.}
We toggle the two mechanisms independently: reweighting in $\{$off, DCCL,
ACCL, DCCL$+$ACCL$\}$ and auxiliary supervision in $\{$off, traffic-light,
stop-line, both$\}$. This gives four headline conditions: \textbf{Baseline}
(both disabled, with weights $\equiv 1$ and aux weights $\lambda^{(k)}_l = 0$,
recovering the standard masked-mean objective); \textbf{BR} (reweighting
only, DCCL$+$ACCL); \textbf{AUX} (rule-token supervision only, with behavioral
reweighting disabled and both
traffic-light and stop-line auxiliary supervision enabled); and
\textbf{Full} (BR$+$AUX). Secondary sweeps vary $w_{\max}$, the per-trigger
and aux weights $\{\lambda^{(k)}_l\}$, the number of reserved tokens, and the
trigger thresholds (set to $-1.5/-3.0/+2.5\,\text{m/s}^2$, $w_{\max}=15$).
\vspace{-1mm}
\paragraph{Trajectory metrics.}
Aggregate displacement error hides tail behavior, so our primary evaluation
is \emph{scenario-bucketed}: a single kinematic mining pass over the
ground-truth trajectory buckets each sample by speed (stationary, low,
medium, high), curvature (low, high), and future kinematic events (future
deceleration, future acceleration, future high curvature), computed
separately on the TL and non-TL slices (Table~\ref{tab:bucketed}). We report
per-bucket ADE at $5$\,s, with future deceleration and future acceleration
as headline numbers, plus overall ADE to confirm the head of the
distribution is not regressed. We follow the open-loop L2/ADE convention of
prior VLA planners but omit open-loop collision rate; its known
insensitivity~\cite{zhai2023admlp,li2024egostatus} partly motivates the
rule-compliance metrics below.

\paragraph{Traffic-light-sliced metrics.}
To isolate rule-sensitive behavior, we partition the validation data using
perception-derived scene properties rather than model predictions. A frame
enters the \emph{TL slice} when the perception stack identifies an ego-relevant
traffic light or qualifying stop line. The remaining eligible frames form the
\emph{non-TL slice}. We report ADE/FDE at $3$\,s for both slices. The non-TL
slice serves as a specificity control for determining whether improvements are
concentrated in rule-governed scenes.
\vspace{-1mm}
\paragraph{Open-loop rule-compliance metrics.}
On the TL validation subset, we report three open-loop rule-compliance
metrics. \emph{Red-light overshoot} measures how often the predicted trajectory
crosses the stop line while the ground truth remains behind it.
\emph{Green-light false stop} measures how often the predicted speed falls below
$1.0$\,m/s while the ground truth proceeds. \emph{Stop-line velocity error}
measures the speed difference at the ground-truth crossing or closest approach
to the stop line. Each metric is computed only on frames that satisfy its
eligibility conditions. These metrics complement displacement error but do not
replace closed-loop evaluation.
\vspace{-1mm}
\paragraph{Auxiliary-task accuracy.}
To check that the reserved tokens actually carry the rule state, we score
the head's predictions against perception targets on held-out frames:
traffic-light presence and color accuracy, stop-line presence accuracy,
distance MAE (m), and orientation MAE ($^{\circ}$). 
\vspace{-1mm}
\paragraph{Comparison to external VLA planners.}
Where possible, we run recent open-source VLA driving models on the same curated training data using model-specific input adapters, then evaluate all
methods on the same curated TL and non-TL slices. This shared data and
evaluation setup enables a direct comparison while retaining each planner's
architecture.
\vspace{-1mm}
\paragraph{Diagnostic: kinematic$\times$rule joint coverage.}
Independent of model quality, we report the fraction of training frames on
which each behavioral trigger fires and the fraction flagged as red-light or
near-stop-line, quantifying the imbalance our method addresses.
\vspace{-1mm}


\vspace{-2mm}
\section{Results}
\label{sec:results}

\begin{figure}[t]
\centering
\begin{minipage}[t]{0.49\linewidth}
\vspace{0pt}
\centering
\includegraphics[width=\linewidth]{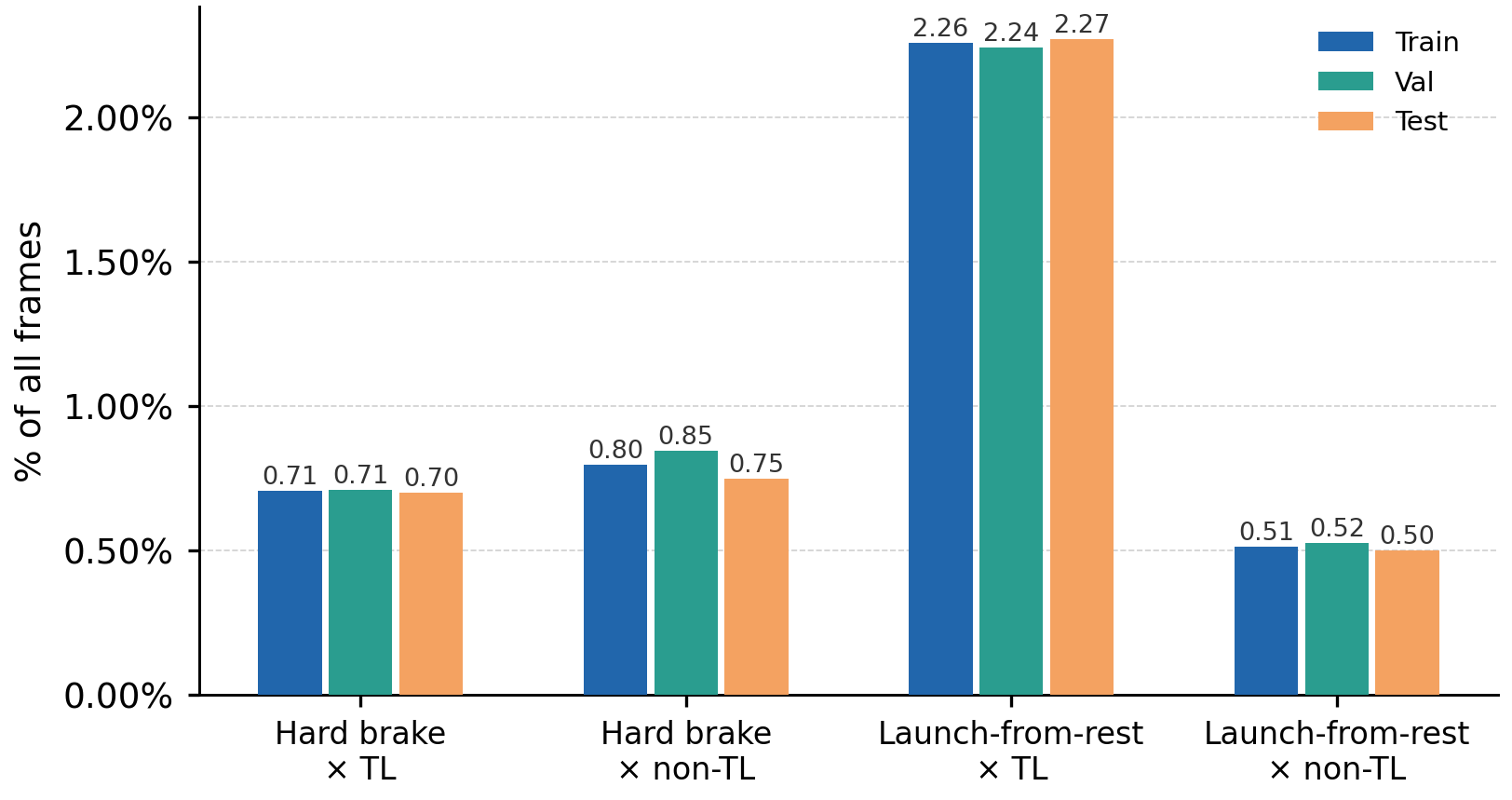}
\captionof{figure}{\textbf{Kinematic event coverage.} Hard-brake and
launch-from-rest frame fractions by split and TL relevance.}
\label{fig:kinematic_coverage}
\end{minipage}\hfill
\begin{minipage}[t]{0.49\linewidth}
\vspace{0pt}
\centering
\scriptsize
\setlength{\tabcolsep}{3pt}
\resizebox{\linewidth}{!}{%
\begin{tabular}{lcccc}
\toprule
 & \multicolumn{2}{c}{TL slice} & \multicolumn{2}{c}{non-TL slice} \\
\cmidrule(lr){2-3}\cmidrule(lr){4-5}
Method & ADE@3s & FDE@3s & ADE@3s & FDE@3s \\
\midrule
SimLingo~\cite{renz2025simlingo}    & 0.309 & 1.071 & 0.301 & 1.056 \\
ORION~\cite{fu2025orion}            & 0.324 & 1.118 & 0.315 & 1.101 \\
Alpamayo-R1~\cite{wang2025alpamayo} & 0.282 & 0.989 & 0.263 & 0.948 \\
\midrule
Baseline & 0.274 & 0.964 & 0.268 & 0.956 \\
BR       & 0.264 & 0.945 & 0.259 & 0.938 \\
AUX      & 0.255 & 0.932 & 0.252 & 0.906 \\
Full     & \textbf{0.247} & \textbf{0.897} & \textbf{0.241} & \textbf{0.876} \\
\bottomrule
\end{tabular}%
}
\setcounter{table}{2}
\captionof{table}{\textbf{TL/non-TL displacement.} ADE/FDE (m, $\downarrow$)
at $3$\,s. All methods share training data and evaluation population;
controlled variants share a backbone, while comparison planners retain their
architectures.}
\label{tab:tl_slice}
\setcounter{table}{1}
\end{minipage}
\end{figure}

\paragraph{Kinematic event coverage.}
Figure~\ref{fig:kinematic_coverage} quantifies the imbalance the method
targets: hard braking covers about $1.5\%$ of frames and launch-from-rest
about $2.8\%$, so both targeted behaviors lie in the long tail.

\paragraph{Traffic-light-sliced displacement.}
Table~\ref{tab:tl_slice} reports the controlled comparison: four conditions
with the same backbone, training data, and evaluation population. Full
attains the lowest error in all four cells, improving TL ADE/FDE at $3$\,s
from $0.274/0.964$ to $0.247/0.897$\,m ($9.9\%/7.0\%$ relative) and non-TL
ADE/FDE from $0.268/0.956$ to $0.241/0.876$\,m ($10.1\%/8.4\%$). Each
single mechanism also improves over Baseline (BR: $0.264/0.945$ TL; AUX:
$0.255/0.932$ TL). The improvements are, however, of similar magnitude on
both slices: the displacement results do not show a TL-specific
concentration of the gain. The rule-compliance metrics in
Table~\ref{tab:rulecomp}, rather than sliced displacement, separate behavior at
signalized intersections.

\paragraph{Open-loop rule compliance.}
Table~\ref{tab:rulecomp} evaluates the joined TL validation subset defined
in Section~\ref{sec:protocol}; each metric uses its own eligibility
predicate and denominator, and we report point estimates. Relative to
Baseline, both AUX and Full reduce red-light overshoot from $7.3\%$ to
$6.8\%$, and Full reduces stop-line velocity error from $0.816$ to
$0.712$\,m/s ($12.7\%$ relative). This comes with a green-light tradeoff:
false stops rise from $3.2\%$ to $4.0\%$ for AUX and $3.9\%$ for Full.
Among the evaluated conditions, Full matches AUX on red-light overshoot and
improves on it for both green-light false stops and stop-line velocity
error.

\begin{figure}[t]
\centering
\begin{minipage}[t]{0.47\linewidth}
\vspace{0pt}
\centering
\scriptsize
\setlength{\tabcolsep}{3pt}
\resizebox{\linewidth}{!}{%
\begin{tabular}{lccc}
\toprule
Method & Red over. (\%) & Green stop (\%) & SL err. (m/s) \\
\midrule
Baseline & 7.3 & 3.2 & 0.816 \\
AUX      & 6.8 & 4.0 & 0.773 \\
Full     & \textbf{6.8} & \textbf{3.9} & \textbf{0.712} \\
\bottomrule
\end{tabular}%
}
\captionof{table}{\textbf{Open-loop rule compliance.} TL-validation point
estimates ($\downarrow$); each metric uses its Section~\ref{sec:protocol}
eligibility population.}
\label{tab:rulecomp}
\end{minipage}\hfill
\begin{minipage}[t]{0.51\linewidth}
\vspace{0pt}
\centering
\scriptsize
\resizebox{\linewidth}{!}{%
\begin{tabular}{lcccc}
\toprule
Method & TL pres. & TL color & SL $\rho$ MAE & SL ori. MAE \\
 & acc. & acc. & (m) & ($^{\circ}$) \\
\midrule
AUX  & 0.9530 & 0.9367 & 2.446 & 6.175 \\
Full & 0.9515 & 0.9366 & 2.626 & 6.493 \\
\bottomrule
\end{tabular}%
}
\setcounter{table}{4}
\captionof{table}{\textbf{Auxiliary-target agreement.} Held-out agreement
with perception targets: accuracies ($\uparrow$) and MAEs ($\downarrow$), not
independent accuracy or causal use.}
\label{tab:auxacc}
\setcounter{table}{3}
\end{minipage}
\end{figure}

\paragraph{Do the mechanisms interact?}
On the evaluated conditions, combining the mechanisms helps: Full improves
over both BR and AUX in every displacement cell of
Table~\ref{tab:tl_slice}, and over AUX on all three rule-compliance metrics
of Table~\ref{tab:rulecomp}. We therefore find that behavioral emphasis and
explicit rule-state representation combine favorably, while noting that our
evidence is at the level of point estimates and that BR's standalone
rule-compliance behavior was not evaluated.

\begin{figure}[t]
\centering
\begin{minipage}[t]{0.49\linewidth}
\vspace{0pt}
\centering
\includegraphics[width=\linewidth]{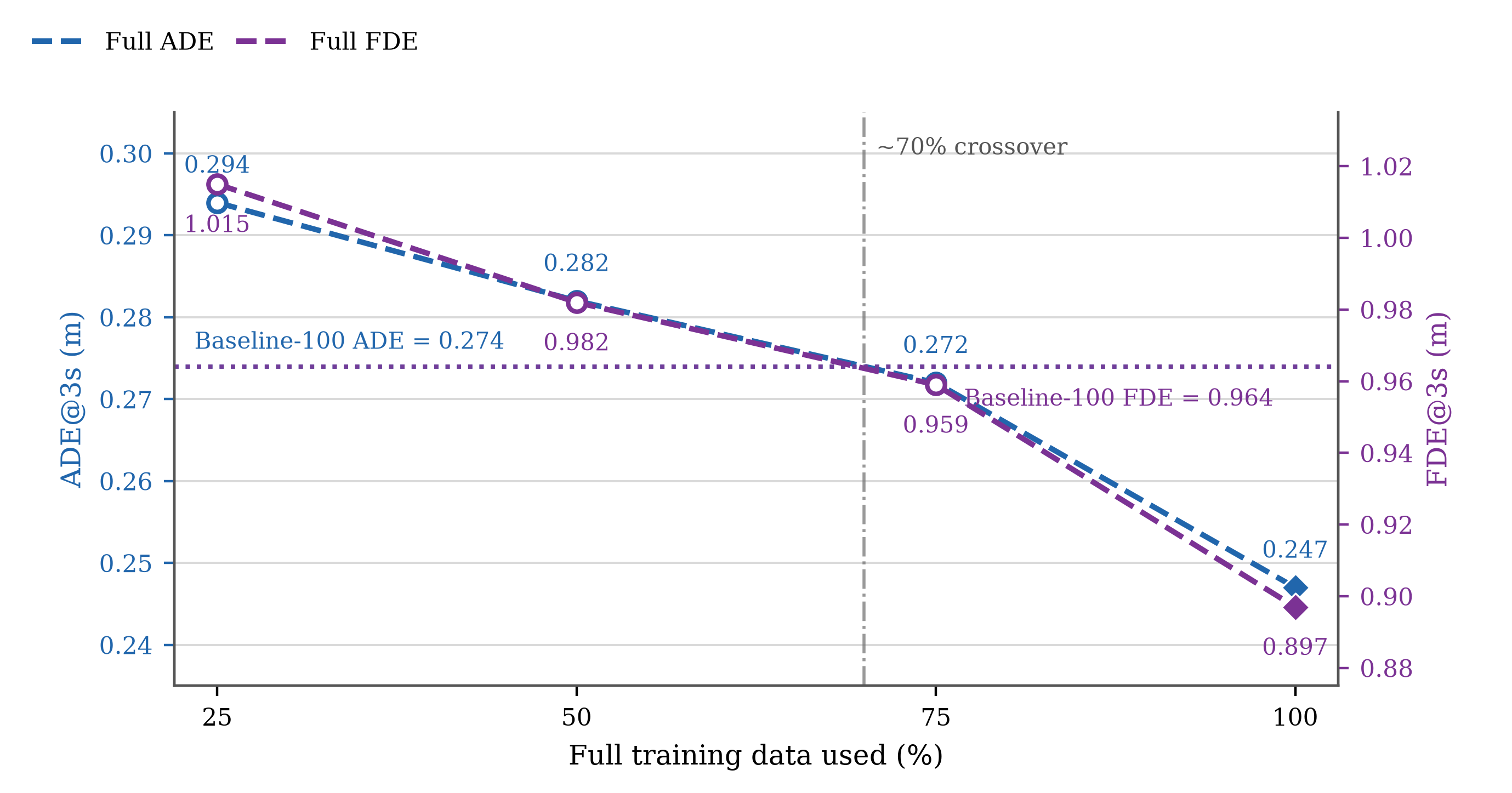}
\end{minipage}\hfill
\begin{minipage}[t]{0.49\linewidth}
\vspace{0pt}
\centering
\includegraphics[width=\linewidth]{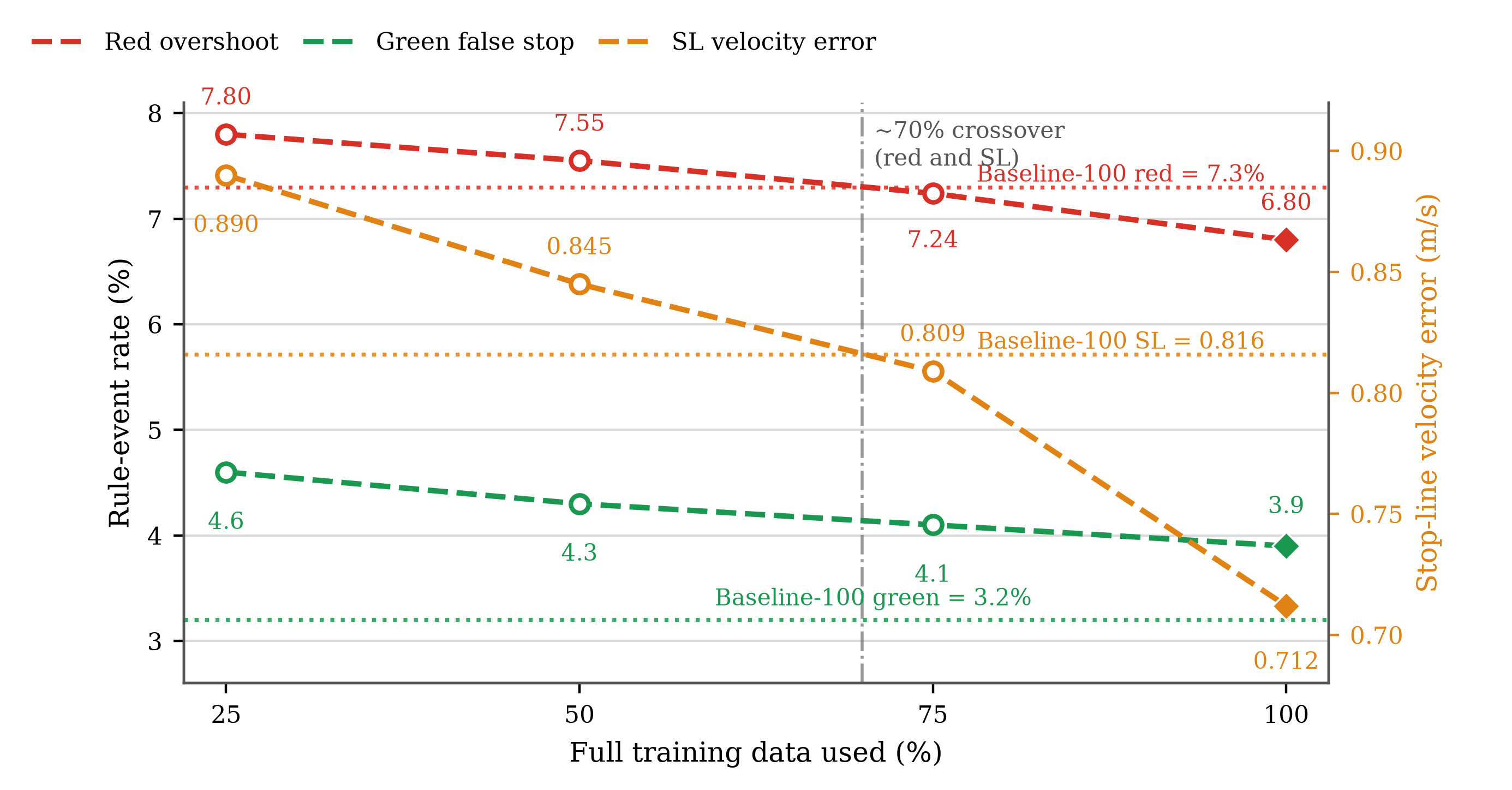}
\end{minipage}
\caption{\textbf{Data-scaling ablation.} Full versus the full-data Baseline
for displacement (left) and rule compliance (right) as the retained training
fraction increases; lower is better.}
\label{fig:data_scaling}
\end{figure}

\paragraph{Training-data scaling.}
Figure~\ref{fig:data_scaling} examines whether targeted supervision can reach
the full-data Baseline with fewer training sequences. The plotted Full trend
crosses Baseline near $70\%$ data for ADE/FDE, red-light overshoot, and
stop-line velocity error. At $75\%$, these four metrics are lower despite
using $25\%$ fewer sequences; green-light false stops remain above Baseline,
so the scaling gain does not remove that tradeoff.

\paragraph{Planner comparison.}
SimLingo, ORION, and Alpamayo-R1 are trained on the same curated dataset as
our models and evaluated on the same TL and non-TL populations. Their results
therefore reflect differences between the planners rather than a dataset or
evaluation-domain mismatch. Full attains the lowest error in all four cells,
while Alpamayo-R1 is the strongest comparison planner.

\paragraph{Scenario-bucketed accuracy.}
Table~\ref{tab:bucketed} slices the same evaluation by ground-truth
scenario. Full reduces overall ADE by $9.8\%$ on the TL slice
($0.7286\!\rightarrow\!0.6573$\,m) and $10.2\%$ on the non-TL slice
($0.6947\!\rightarrow\!0.6240$\,m), confirming that aggregate accuracy is
not regressed, and is lower in 19 of 20 slice--scenario cells. The buckets
the method targets improve most: future-deceleration ADE drops by
$0.1187/0.1349$\,m and future-acceleration ADE by $0.0708/0.1354$\,m on the
TL/non-TL slices. The one non-improving cell, stationary TL frames
($+0.0016$\,m), is consistent with a policy that already predicts near-zero
motion when stopped.

\begin{table}[t]
\centering
\caption{\textbf{Scenario-bucketed ADE at $5$\,s on the curated validation
extract} (m, $\downarrow$). The TL slice contains frames with an
ego-relevant traffic light or stop line; the non-TL slice
contains the remaining eligible frames. Full has lower
point-estimate ADE than Base in 19 of 20 slice--scenario cells.}
\label{tab:bucketed}
\scriptsize
\setlength{\tabcolsep}{4pt}
\resizebox{\linewidth}{!}{%
\begin{tabular}{lrrrrrr}
\toprule
& \multicolumn{3}{c}{TL slice } & \multicolumn{3}{c}{non-TL slice } \\
\cmidrule(lr){2-4}\cmidrule(lr){5-7}
Scenario & Base & Full & $\Delta$Full & Base & Full & $\Delta$Full \\
\midrule
Stationary            & 0.3361 & 0.3377 & +0.0016 & 0.4300 & 0.3645 & -0.0655 \\
Low speed             & 0.5942 & 0.5550 & -0.0392 & 0.8185 & 0.7234 & -0.0951 \\
Medium speed          & 0.8302 & 0.7318 & -0.0984 & 0.7861 & 0.6821 & -0.1040 \\
High speed            & 0.7154 & 0.6667 & -0.0487 & 0.5664 & 0.5356 & -0.0308 \\
Low curvature         & 0.6397 & 0.5825 & -0.0572 & 0.6131 & 0.5598 & -0.0533 \\
High curvature        & 0.9670 & 0.8608 & -0.1062 & 1.0115 & 0.8764 & -0.1351 \\
Future deceleration   & 0.9775 & 0.8588 & \textbf{-0.1187} & 1.0642 & 0.9293 & \textbf{-0.1349} \\
Future acceleration   & 1.0937 & 1.0229 & -0.0708 & 1.1473 & 1.0119 & \textbf{-0.1354} \\
Future high curvature & 1.2693 & 1.1741 & -0.0952 & 1.3922 & 1.2379 & \textbf{-0.1543} \\
\midrule
\textbf{Overall}      & \textbf{0.7286} & \textbf{0.6573} & \textbf{-0.0713} & \textbf{0.6947} & \textbf{0.6240} & \textbf{-0.0707} \\
\bottomrule
\end{tabular}%
}
\vspace{2pt}
\begin{minipage}{0.98\linewidth}
\footnotesize Base denotes the VLA baseline with both BR and AUX disabled.
$\Delta$Full is Full$-$Base.
\end{minipage}
\end{table}

\paragraph{Auxiliary-target agreement.}
Table~\ref{tab:auxacc} verifies that the reserved tokens carry the rule
state: traffic-light presence and color agreement reach $95.3\%$ and
$93.7\%$, and stop-line distance MAE is under $2.7$\,m. Adding BR (Full)
slightly reduces auxiliary agreement relative to AUX alone while improving
trajectory metrics, consistent with the reweighted objective shifting
emphasis toward the trajectory terms. These metrics measure agreement with
perception-derived targets, not causal use by the trajectory decoder.

\begin{figure}[t]
\centering
\setlength{\tabcolsep}{1.5pt}
\renewcommand{\arraystretch}{1.2}
\begin{tabular}{@{}c@{\hspace{2pt}}ccc@{}}
\rotatebox{90}{\makebox[0.169\linewidth][c]{\scriptsize\textbf{Stop for red}}} &
\includegraphics[width=0.31\linewidth]{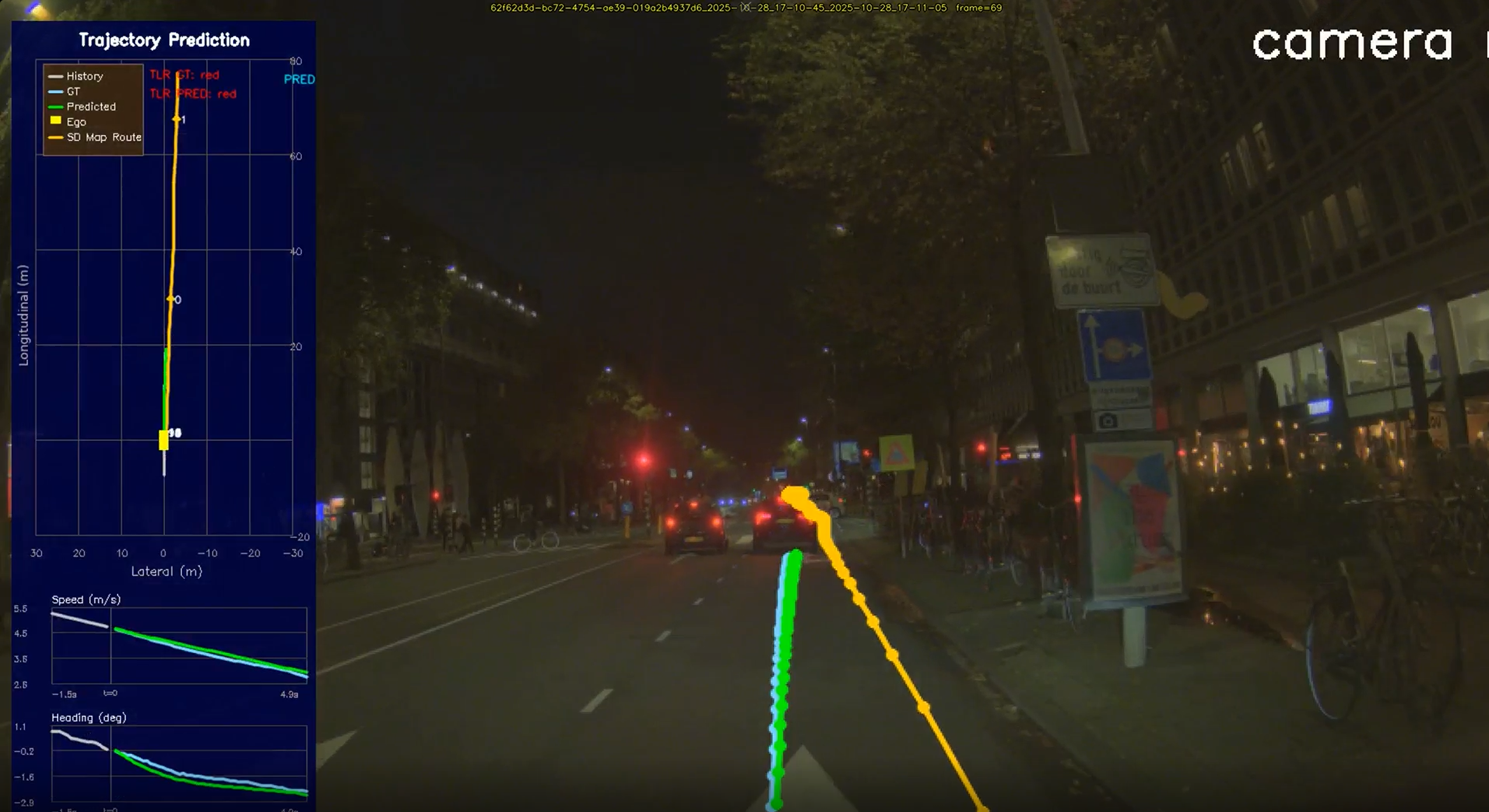} &
\includegraphics[width=0.31\linewidth]{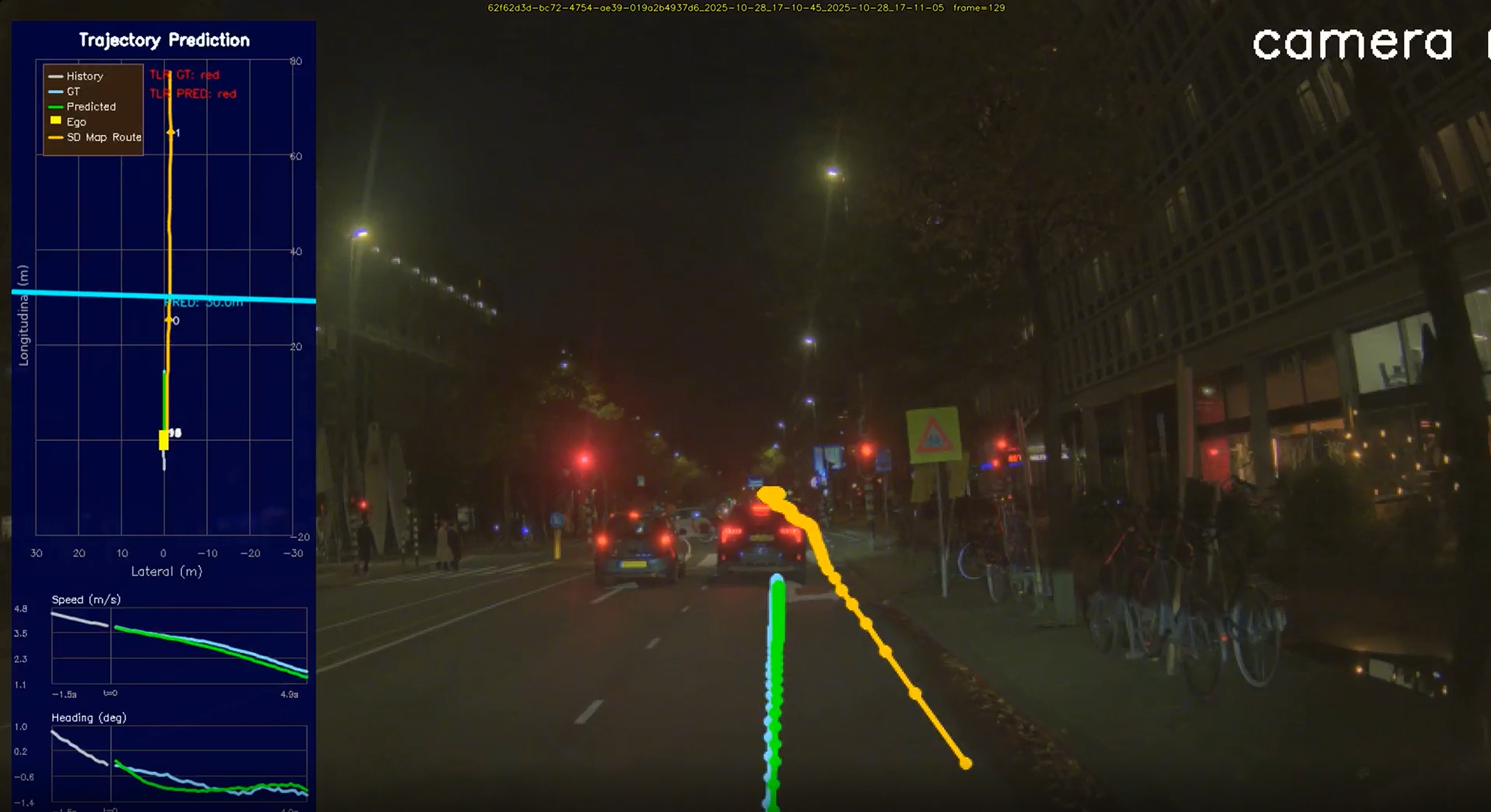} &
\includegraphics[width=0.31\linewidth]{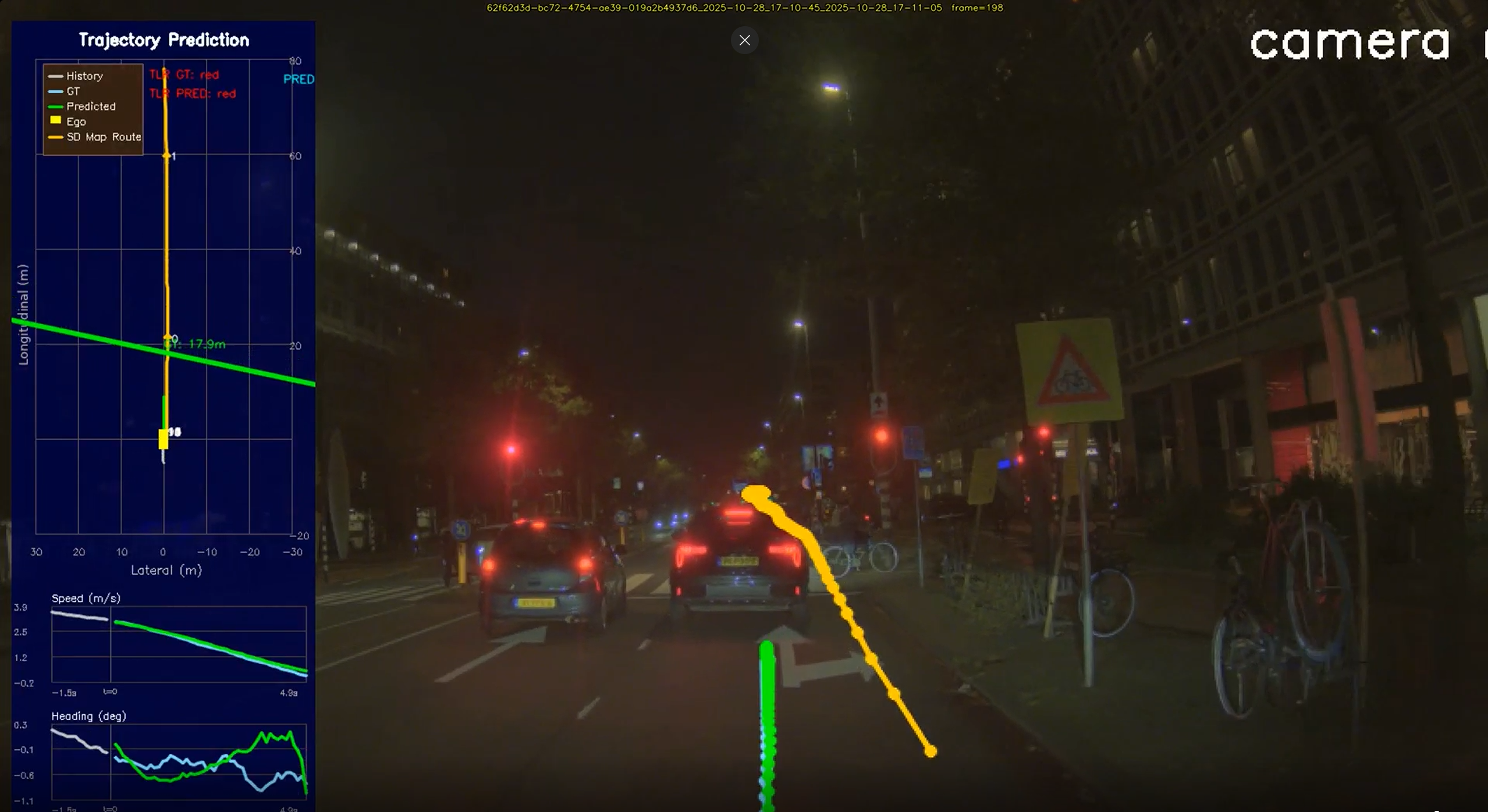} \\[1pt]
\rotatebox{90}{\makebox[0.169\linewidth][c]{\scriptsize\textbf{Red\,$\rightarrow$\,green}}} &
\includegraphics[width=0.31\linewidth]{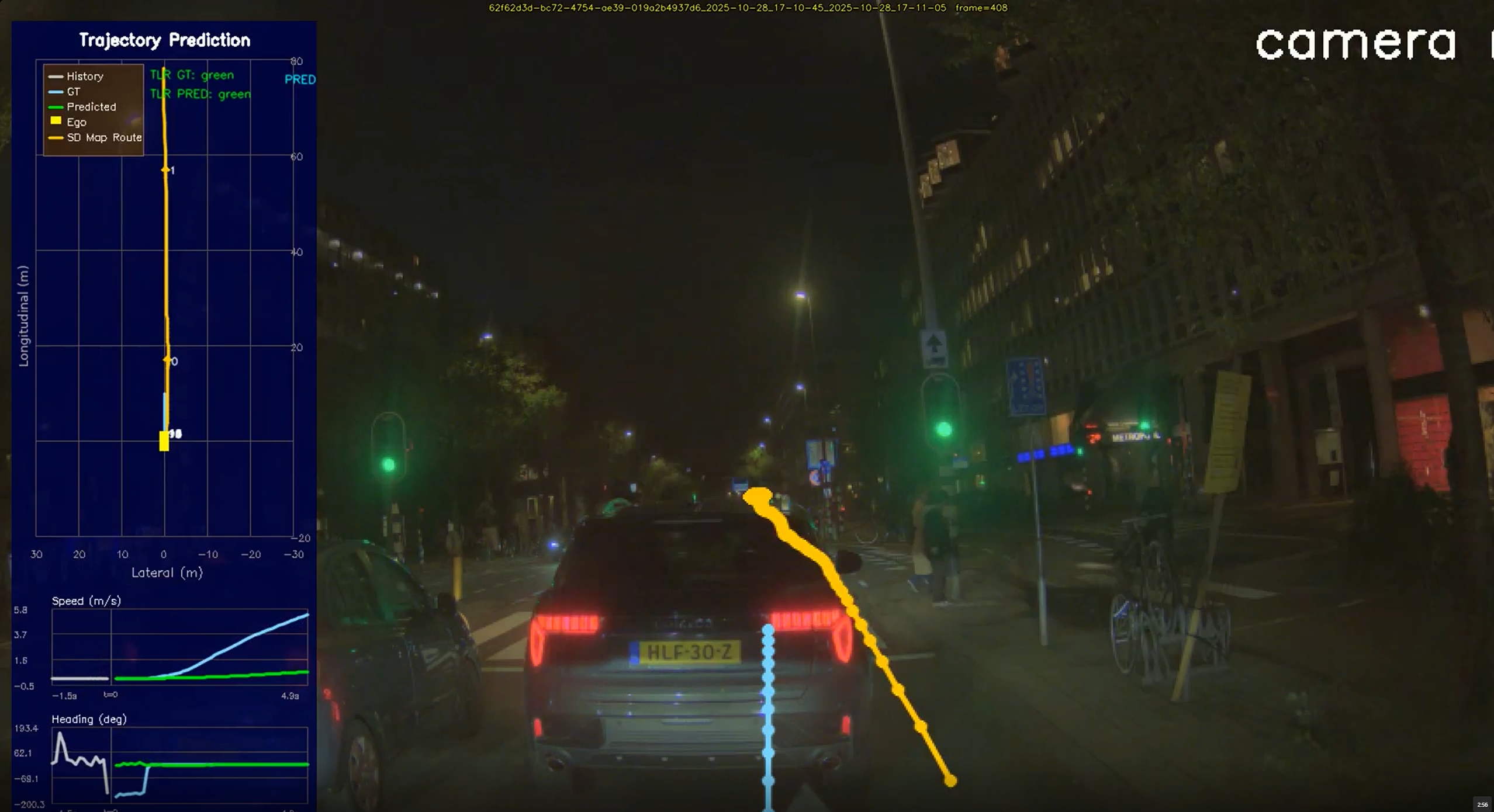} &
\includegraphics[width=0.31\linewidth]{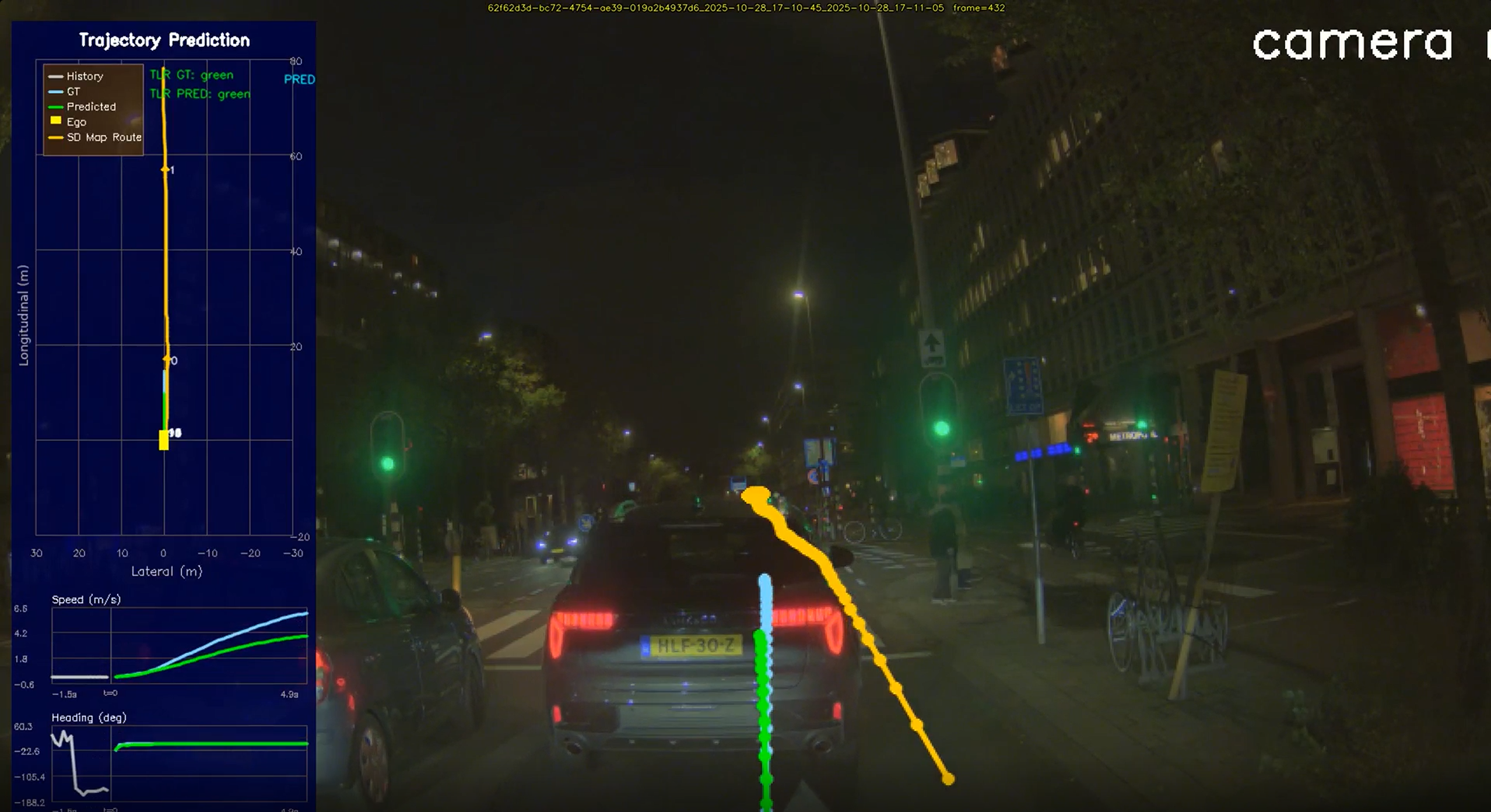} &
\includegraphics[width=0.31\linewidth]{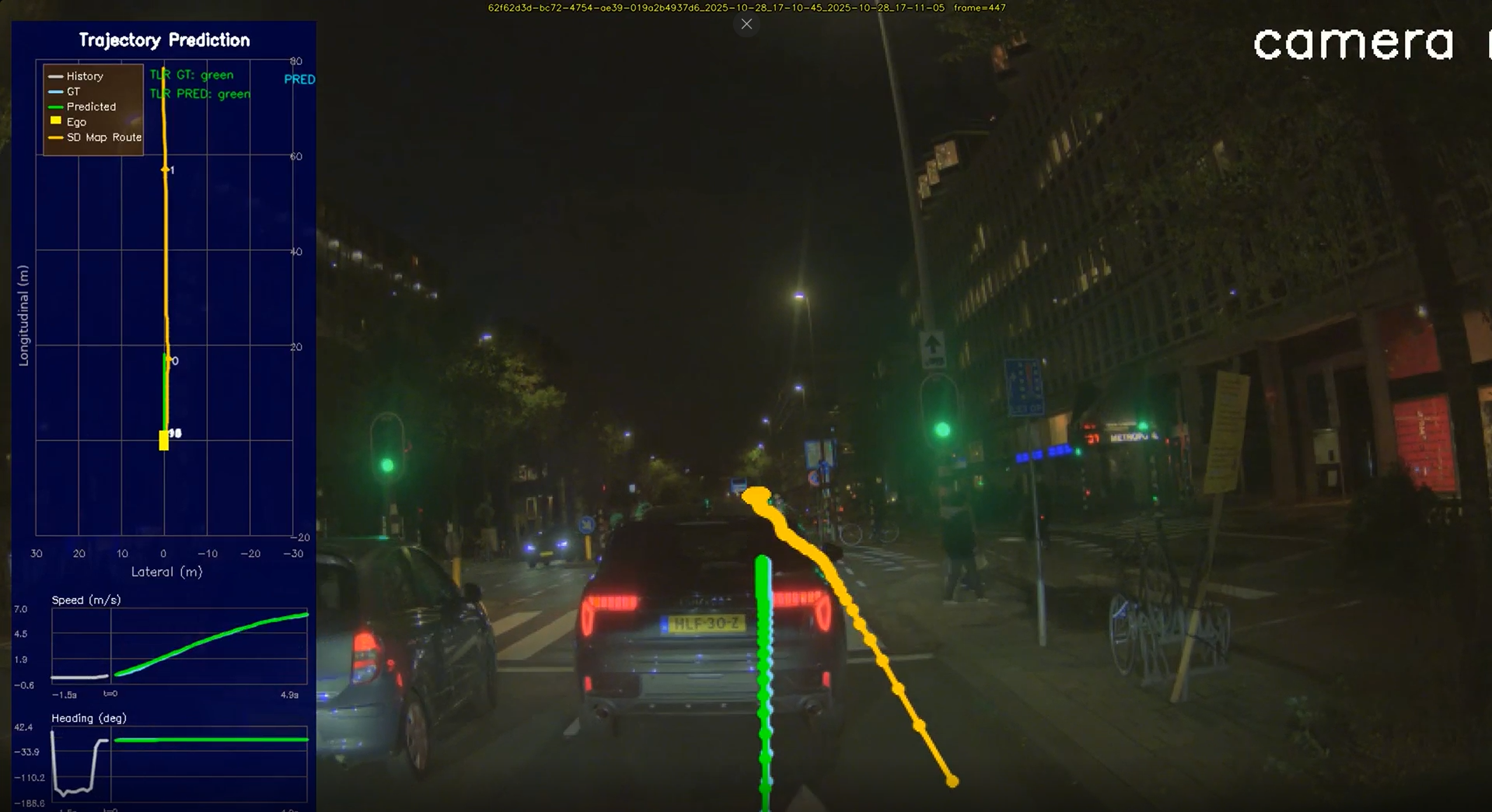} \\
\end{tabular}
\caption{\textbf{Qualitative red-to-green sequence.} Predictions (green) stop at red (top) and accelerate after green (bottom), compared with reference futures (blue). Insets show speed and TLR GT/PRED; this example is illustrative, not causal evidence.}
\label{fig:overlays}
\end{figure}

\paragraph{Qualitative illustration.}
Figure~\ref{fig:overlays} shows selected frames from one
signalized-intersection sequence, illustrating agreement and disagreement
between the predicted and reference futures. It is not used to infer a
causal effect of the auxiliary representation.

\section{Discussion and Limitations}
\label{sec:discussion}

\paragraph{Interpreting the results.}
Full's displacement gains are of similar magnitude on the TL and non-TL
slices, so the separation between conditions at signalized intersections
comes from the rule-compliance metrics, not sliced displacement. Those
metrics show an asymmetric tradeoff: AUX and Full reduce red-light overshoot
but increase green-light false stops, indicating a conservative bias. Because
AUX has the largest increase, the tradeoff cannot be attributed solely to
braking reweighting. Full mitigates the increase relative to AUX, consistent
with its launch emphasis, although the absence of BR standalone rule metrics
and uncertainty estimates precludes causal attribution. A rule-conditioned
weighter that emphasizes launches under green is a natural next step.
\vspace{-2mm} 
\paragraph{Limitations.}
(i)~\textbf{Longitudinal-only triggers}; lateral rarity is not directly targeted. (ii)~\textbf{Handcrafted thresholds}, set to physically interpretable
defaults that each fleet may need to tune.
(iii)~\textbf{No coverage guarantee}: a behavior missing from the logs
cannot be produced, so the method complements data curation rather than
replacing it.
(iv)~\textbf{Perception dependence}: auxiliary targets inherit upstream
perception errors, and auxiliary agreement is measured against those
targets, not independent ground truth.
(v)~\textbf{Evaluation scope}: we report point estimates without
uncertainty intervals, do not evaluate BR's standalone rule compliance,
and do not quantify the comfort impact of reweighting.
(vi)~\textbf{Open-loop evaluation only}; closed-loop scoring on real
scenes would require rendering off the logged path.
(vii)~\textbf{Single curated dataset}; transfer to public benchmarks
(nuPlan, NAVSIM) is left to future work.


\section{Conclusion}
\label{sec:conclusion}
We presented an approach to the driving long tail that uses automatically
generated training signals without additional manual rule annotation.
Behavioral reweighting uses the expert future's longitudinal
dynamics to emphasize rare braking and launching frames through a
scale-preserving reduction that recovers the baseline objective exactly
when disabled. Auxiliary rule-token supervision reads perception-derived
traffic-light and stop-line state from reserved tokens after
language-model fusion, leaving the trajectory head unchanged. On a
real-world driving-log dataset, the combined model reduced red-light
stop-line overshoot from $7.3\%$ to $6.8\%$, reduced stop-line velocity
error by $12.7\%$, and achieved the lowest traffic-light-sliced ADE/FDE
among all evaluated conditions, improving over each mechanism alone, at
the cost of increased green-light false stops, which Full mitigates relative
to AUX. Both
mechanisms preserve the existing trajectory decoder and introduce no
autoregressive language-generation stage. Future work includes
closed-loop evaluation, uncertainty quantification, rule-conditioned
weighting to address the conservatism tradeoff, and transfer to public
benchmarks.



\IfFileExists{eccv-paper/splncs04.bst}{%
  \bibliographystyle{eccv-paper/splncs04}%
}{%
  \bibliographystyle{splncs04}%
}
\IfFileExists{eccv-paper/references.bib}{%
  \bibliography{eccv-paper/references}%
}{%
  \bibliography{references}%
}
\end{document}